\documentclass{spie}

\usepackage{amsmath,amsfonts,amssymb}
\usepackage{graphicx}
\usepackage[colorlinks=true, allcolors=blue]{hyperref}

\usepackage{multirow}
\usepackage{multicol}

\usepackage{comment}

\RequirePackage{xspace}
\makeatletter
\DeclareRobustCommand\onedot{\futurelet\@let@token\@onedot}
\def\@onedot{\ifx\@let@token.\else.\null\fi\xspace}
\def\etal{\emph{et al}\onedot}
\makeatother

\title{M3DDM+: An improved video outpainting\\ by a modified masking strategy}

\author{Takuya Murakawa${}^1$}
\author{Takumi Fukuzawa}
\author{Ning Ding}
\author{Toru Tamaki${}^2$}
\affil{Nagoya Institute of Technology}

\authorinfo{
${}^1$t.murakawa.080@nitech.jp, 
${}^2$tamaki.toru@nitech.ac.jp\\
}

\begin{document} 
\maketitle

\begin{abstract}
M3DDM provides a computationally efficient framework for video outpainting via latent diffusion modeling. However, it exhibits significant quality degradation --- manifested as spatial blur and temporal inconsistency --- under challenging scenarios characterized by limited camera motion or large outpainting regions, where inter-frame information is limited.
We identify the cause as a training-inference mismatch in the masking strategy: M3DDM's training applies random mask directions and widths across frames, whereas inference requires consistent directional outpainting throughout the video. To address this, we propose \emph{M3DDM+}, which applies uniform mask direction and width across all frames during training, followed by fine-tuning of the pretrained M3DDM model. Experiments demonstrate that M3DDM+ substantially improves visual fidelity and temporal coherence in information-limited scenarios while maintaining computational efficiency. The code is available at \url{https://github.com/tamaki-lab/M3DDM-Plus}.
\end{abstract}


\begin{figure}[h]
    \centering
    \includegraphics[width=1\linewidth]{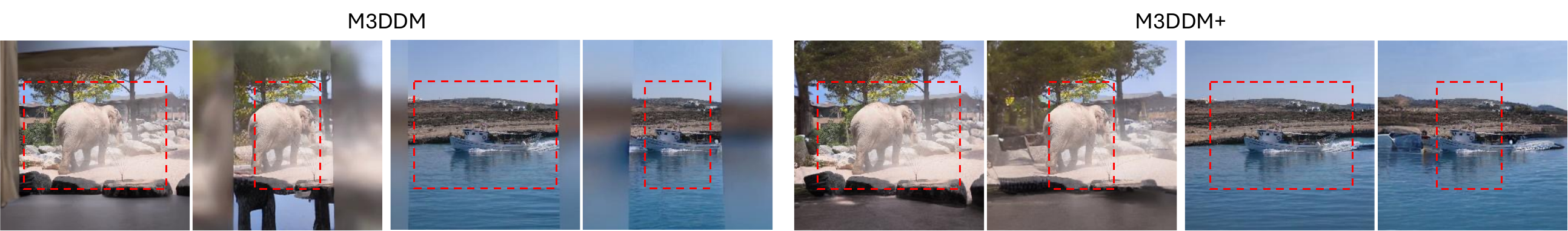}
    \caption{Comparison of M3DDM and our \emph{M3DDM+}.
    Red boxes indicate the boundary between the input frame regions and the generated regions.}
    \label{fig:intro_comparison}
\end{figure}

\begin{figure}[t]
    \centering
    \includegraphics[width=1\linewidth]{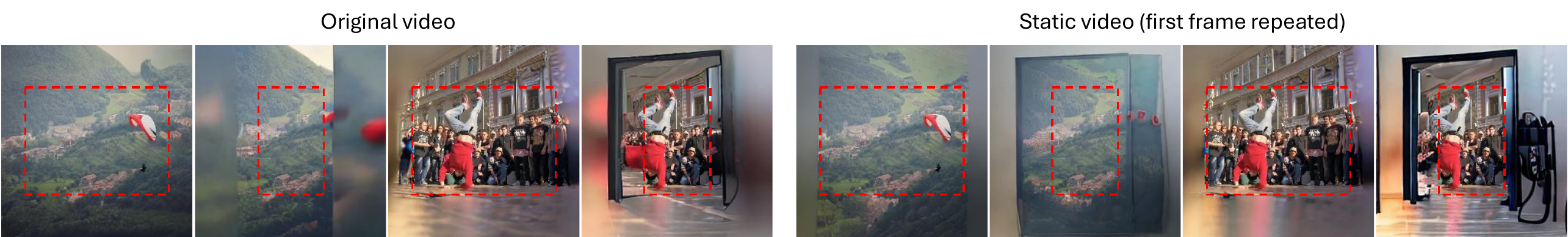}
    \caption{Failure examples of M3DDM generation. Red boxes indicate the boundary between the input frame regions and the generated regions. In this experiment, 25\% or 66\% of the lateral margins of a 16:9 input video are cropped, and video outpainting is performed to restore the original 16:9 aspect ratio. The left two columns present first frames of generated videos under a standard outpainting setting with camera motion, while the right two columns illustrate outpainting results for static videos constructed by replacing all frames of the original video with its first frame.}
    \label{fig:M3DDM_failure}
\end{figure}

\section{INTRODUCTION}
\label{sec:intro}

Video outpainting is the task of generating out-of-frame regions to transform an input video into an arbitrary aspect ratio while maintaining spatiotemporal consistency by leveraging in-frame and inter-frame visual information. This capability addresses a practical challenge: as vertical videos (9:16 aspect ratio) captured on smartphones have become increasingly common, they create prominent black bars on the left and right sides of the screen when displayed on conventional horizontal displays (16:9 or wider), such as televisions and computer monitors. By synthesizing coherent content in these void regions, video outpainting enables more immersive viewing experiences across diverse display configurations.

In image outpainting, recent advances in GANs \cite{goodfellow_NIPS2014_GANs} and diffusion models \cite{ho_NEURIPS2020_diffusion, Dickstein_PMLR2015_diffusion} have achieved high-quality, high-resolution synthesis with strong spatial coherence. Video outpainting, however, presents a fundamentally more challenging problem: in addition to spatial consistency, it must maintain temporal coherence across object motion, human actions, and camera dynamics throughout the sequence.

Among existing video outpainting methods, M3DDM \cite{Fan_ACMMM2023_M3DDM} preserves temporal consistency by employing a 3D U-Net \cite{ozgun_MICCAI2016_3D_U-Net} architecture for temporal modeling and processing uniformly sampled frames across the input sequence. More recent approaches refine this paradigm in different directions. 
MOTIA~\cite{wang_ECCV2024_MOTIA} focuses on improving generation quality without retraining on large-scale video datasets by performing input-specific adaptation: it inserts LoRAs into a pretrained diffusion model and conducts pseudo outpainting learning on each source video, then performs pattern-aware outpainting that combines the learned video-specific patterns with the model's generative prior.
In practice, MOTIA still introduces additional per-video computation due to this adaptation stage.
In contrast, Infinite-Canvas \cite{chen_AAAI2025_Infinite-Canvas} (a.k.a. Follow-Your-Canvas) targets high-resolution and large-area outpainting by distributing generation across spatial windows and seamlessly merging them, while injecting the source video and its relative positional relations into each window to enforce layout coherence and spatiotemporal consistency. 
Although these methods achieve superior visual quality and support higher output resolutions, they come with substantial computational cost: per-video adaptation increases inference time in MOTIA, and Infinite-Canvas enables expansion from $512 \times 512$ to $1152 \times 2048$ resolution but requires a GPU with at least 60\,GB of memory for both training and inference.
Although these approaches achieve high-resolution, high-quality generation at the expense of substantial computational resources, practical deployment scenarios require approaches that maximize generation quality under constrained computational budgets.

M3DDM, a computationally efficient approach, achieves lightweight video outpainting by encoding input videos into a latent space via a VAE encoder and employing a latent diffusion model that performs masking and denoising operations in the latent representation.
However, recent work \cite{zhong_arXiv2025_OutDreamer, chen_AAAI2025_Infinite-Canvas, wang_ECCV2024_MOTIA} has identified that M3DDM's generated outputs tend to exhibit blurriness and are susceptible to temporal inconsistencies and fine-grained detail degradation across frames.
As illustrated in Fig.~\ref{fig:M3DDM_failure}, this quality degradation becomes particularly severe under conditions where out-of-frame information in the input video is limited, such as when the outpainting region is relatively large compared to the visible input area (second and fourth panel) or when camera motion is constrained (third and fourth panels).
The fundamental cause of this quality degradation can be attributed to a structural inconsistency in M3DDM's training strategy. During training, M3DDM applies independent random mask patterns to each frame, resulting in different visible and generated regions between frames, even within the same training video.
This design is fundamentally misaligned with the typical inference scenario --- namely, generating consistent outpainting regions with unified direction and width across all frames of the input video.
Due to this mismatch of training and inference conditions, the model acquires a tendency to excessively rely on information obtained from neighboring frames during the training process.
Consequently, under conditions where visual information shared across frames is limited during inference, the model's generative capability significantly deteriorates, resulting in blurred outputs accompanied by spatial detail loss and temporal inconsistencies.

In this work, we identify and address the fundamental cause of quality degradation in M3DDM --- the mismatch of training and inference distributions --- by proposing \emph{M3DDM+}, an improved framework with a strategically optimized masking scheme (Fig.~\ref{fig:intro_comparison}).
M3DDM+ enforces spatially consistent mask directions and widths across all frames within each training video, thereby bridging the gap between training and inference conditions.
This training strategy refinement mitigates the model's over-reliance on cross-frame contextual cues from adjacent frames, enabling robust generation of high-fidelity, temporally coherent videos even when out-of-frame spatial information is substantially limited.
Our approach yields substantial improvements in both perceptual quality and spatiotemporal consistency while fully retaining the computational efficiency of the original M3DDM framework.

\section{RELATED WORK}
Image outpainting has been extensively studied through GAN-based approaches
\cite{sabini_arXiv2018_PaintingOutsidetheBox, cheng_CVPR2022_InOut, Xu_ACCV2022_DHGGAN, vanhoorick_2020arXiv_ImageOutpaintingandHarmonizationusingGenerativeAdversarialNetworks, Lin_CVPRW2021_EdgeGuidedProgressivelyGenerativeImageOutpainting}
and diffusion-based methods
\cite{zhang_ICLR2024_continuousmultiple, song_ICCV2025_progressive, Tsai_ICCV2025_LightsOut, yang_ACCV2024_vip}.
In contrast, video outpainting remains significantly more challenging due to the necessity of maintaining not only spatial consistency within each frame but also temporal coherence across frames --- including object motion, human actions, and camera movements --- while incurring substantially higher computational costs during both training and inference. Consequently, relatively few methods have been proposed for this task.

As an early seminal work in video outpainting, Dehan \etal \cite{Dehan_CVPRW2022_Dehan} specifically addressed the background region outpainting, while explicitly excluding foreground elements.
Their pipeline first applies video object segmentation and video inpainting to remove dynamic foreground objects, producing background-only video sequences. These sequences are then expanded using optical flow and image outpainting to synthesize the outpainted regions while preserving temporal consistency across frames.

M3DDM \cite{Fan_ACMMM2023_M3DDM} extends a pretrained image-based latent diffusion model to the temporal dimension and performs additional training for outpainting tasks, allowing video outpainting that accommodates dynamic objects.
Specifically, M3DDM replaces the backbone U-Net \cite{Olaf_MICCAI2015_U-Net} of the base latent diffusion model with a 3D U-Net, allowing the model to leverage temporal information from adjacent frames for video generation.
M3DDM is then trained on a large-scale video dataset with masks applied to between one and four of the borders (top, bottom, left, or right), enabling robust outpainting capabilities.
During inference, frames uniformly sampled from the entire input video are fed into the 3D U-Net's cross-attention mechanism, facilitating the incorporation of out-of-frame contextual information from the full video sequence. Additionally, a coarse-to-fine inference pipeline is introduced to mitigate artifact accumulation during long video generation.
Despite these contributions, M3DDM exhibits notable failure modes, including blurred outputs and temporal instability, as illustrated in Fig.~\ref{fig:M3DDM_failure}. These limitations have been reported in subsequent works \cite{zhong_arXiv2025_OutDreamer, chen_AAAI2025_Infinite-Canvas, wang_ECCV2024_MOTIA}.

MOTIA \cite{wang_ECCV2024_MOTIA} addresses the blur artifacts observed in M3DDM by attributing them to a domain distribution mismatch between training clips and input videos provided at inference time. To mitigate this domain gap, MOTIA performs lightweight test-time fine-tuning using Low-Rank Adapters (LoRAs) \cite{hu_ICLR2022_LoRA} on a per-video basis.
While M3DDM trains a pretrained model on large-scale datasets for video outpainting, MOTIA freezes the pretrained backbone and combines two key strategies: (1) input-specific adaptation via LoRA fine-tuning tailored to each input video, and (2) pattern-aware outpainting that modulates LoRA influence spatially --- applying stronger adaptation near original frame boundaries to preserve input-specific patterns, while relying more heavily on the pretrained model's generalization capability in distant regions. This spatial modulation effectively reduces blur and enhances output fidelity.
Despite successfully suppressing blur, the test-time adaptation process incurs substantially higher computational overhead compared to Dehan \etal \cite{Dehan_CVPRW2022_Dehan} and M3DDM, limiting its practical applicability in latency-sensitive scenarios.

Recent advances in video outpainting have introduced specialized methods for high-resolution synthesis \cite{chen_AAAI2025_Infinite-Canvas}, architectures based on Diffusion Transformers (DiT) \cite{zhong_arXiv2025_OutDreamer}, and techniques utilizing 3D Gaussian Splatting \cite{yu_CVPR2025_Unboxed}, yielding notable improvements in both generation fidelity and temporal coherence.
However, these enhanced capabilities entail substantial increases in computational overhead and GPU memory usage, thus limiting their applicability in resource-constrained deployment scenarios.

In this work, we focus on M3DDM, a computationally efficient video outpainting framework built on latent diffusion models. We identify the underlying causes of its quality degradation under information-limited conditions and propose a refined training strategy that optimizes directional mask application during fine-tuning, thereby achieving high-fidelity video generation while preserving computational efficiency.

\section{METHOD}

\subsection{Diffusion Model}
\label{sec:DiffusionModel}

Diffusion models \cite{ho_NEURIPS2020_diffusion, Dickstein_PMLR2015_diffusion} are generative probabilistic frameworks that learn data distributions by progressively denoising samples drawn from a Gaussian prior.
The model comprises a forward diffusion process that sequentially corrupts the input data $x_0$ with Gaussian noise $\epsilon$ to produce a sequence $x_0, x_1, \ldots, x_T$, and a reverse denoising process that reconstructs the clean data.
The forward process is defined as
$x_t = \sqrt{\bar{\alpha}_t}\, x_0 + \sqrt{1-\bar{\alpha}_t}\, \epsilon$,
where $\epsilon\sim\mathcal{N}(0,I)$ is a Gaussian noise, and $\bar{\alpha}_t$ denotes a noise schedule hyperparameter.
The reverse process trains a denoiser $\epsilon_\theta$ to predict the noise $\epsilon$ added in the time step $t$, minimizing the mean squared error loss
$
\mathcal{L}_{DM}
= \mathbb{E}_{x, \epsilon, t}
\left[
\| \epsilon - \epsilon_\theta(x_t, t) \|_2^2
\right]$.
Here, $\epsilon_\theta$ is typically instantiated as a 3D U-Net, and $\theta$ denotes the learnable parameters.

M3DDM employs a Latent Diffusion Model (LDM) \cite{Rombach_CVPR2022_SDv1-5}, which significantly reduces computational overhead by performing the diffusion process in a compressed latent space rather than directly in the pixel space.
Specifically, a pretrained Variational Auto-Encoder (VAE) \cite{Kingma_ICLR2014_VAE} encoder $E$ maps the input video $x$ to a latent representation $z = E(x)$, and the model is trained to minimize the loss
$
\mathcal{L}_{LDM}
= \mathbb{E}_{x, \epsilon, t}
\left[
\bigl\| \epsilon - \epsilon_\theta(z_t, t) \bigr\|_2^2
\right]
$ in the latent space.
At inference, the denoised latent representation $\hat{z}_0$ is decoded via the VAE decoder to yield the final output video.

\subsection{Mask Strategy}

\begin{figure}[t]
    \centering
    \includegraphics[width=1\linewidth]{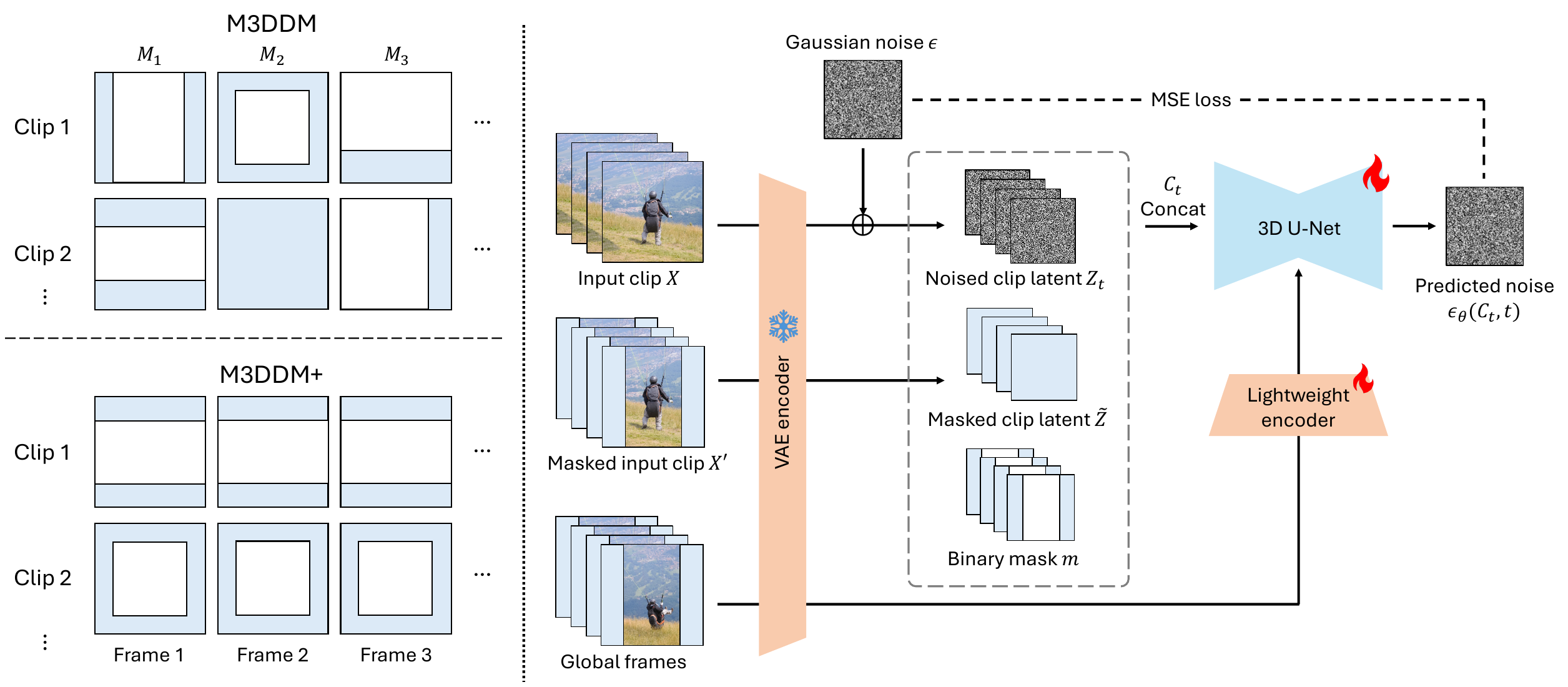}

    \caption{Comparison of mask strategies (left) and the training pipeline of M3DDM and M3DDM+ (right). In the comparison of mask strategy, the blue area indicates mask region and the white area indicates the input clip region. M3DDM applies frame-wise varying mask directions and ratios within each video sequence, whereas our method enforces spatiotemporally consistent masking across all frames. The 3D U-Net receives a channel-wise concatenation: noised latent representations of the input video, latent representations of the masked input, and a binary mask encoding the masked regions.}

    \label{fig:M3DDM_model}
\end{figure}

The left side of Fig.~\ref{fig:M3DDM_model} illustrates the masking strategies of M3DDM and our proposed M3DDM+.
In the M3DDM training procedure, each frame $\tau$ ($\tau=1,\ldots,T$) independently samples a mask $M_\tau$ with randomized direction and ratio.
Specifically, the masking direction is sampled from five strategies
--- four-sided, two-sided (vertical or horizontal), single-sided, random, and full-frame ---
with probabilities 0.2, 0.1, 0.35, 0.1, and 0.25, respectively. The \emph{mask ratio}, uniformly sampled from [0.15, 0.75] (where 1.0 represents the full distance from the frame boundary to the center), determines the spatial extent of the masked region.

In contrast, M3DDM+ aligns the training scenario with the inference scenario by enforcing temporal consistency in masking: all frames within a training video share the same mask direction and ratio, i.e., $M_1 = M_2 = \cdots = M_T$.
This design addresses the training-inference mismatch in M3DDM --- where per-frame random masking during training differs from uniform directional masking at inference. It also encourages the model to synthesize out-of-frame content from intra-frame information alone, even when inter-frame information is unavailable.

\subsection{Training}
The right side of Fig.~\ref{fig:M3DDM_model} illustrates the architecture of M3DDM and M3DDM+.
Given an input video
$X = [ x_1, \ldots, x_T], \ x_\tau \in \mathbb{R}^{H \times W \times 3}$,
we define a binary mask $M_\tau \in \{0,1\}^{H \times W}$ for each frame, where 0 denotes known regions and 1 denotes regions to be outpainted. The masked frames are obtained as
$x'_\tau = x_\tau \odot (1 - M_\tau)$.
Both $x_\tau$ and $x'_\tau$ are encoded using a pretrained VAE encoder $E$, producing latent representations 
$Z = [E(x_1), \ldots, E(x_T)]$ and
$\tilde{Z} = [E(x'_1), \ldots, E(x'_T)]$
of the original and masked videos, where
$Z, \tilde{Z} \in \mathbb{R}^{T \times h \times w \times c}$.

During the diffusion process, following Sect.~\ref{sec:DiffusionModel}, at each training step we sample the time step $t$ and Gaussian noise $\epsilon$, then apply noise to the latent video $Z$ via
$Z_t = \sqrt{\bar{\alpha}_t}\, Z + \sqrt{1 - \bar{\alpha}_t}\, \epsilon$.
The noised latent $Z_t$, the masked latent video $\tilde{Z}$, and the downsampled binary mask $m = [m_1, \ldots, m_T]$ (which is a sequence of $M_\tau$ resized to match the VAE stride) are concatenated channel-wise to form
$C_t = \mathrm{concat}(Z_t, \tilde{Z}, m)$,
which is fed into the 3D U-Net.
The denoiser $\epsilon_\theta$ is then trained to predict $\epsilon$ from $C_t$ by minimizing
$
\mathcal{L}_{\mathrm{LDM}}
= \mathbb{E}_{X, \epsilon, t}
\bigl[
\| \epsilon - \epsilon_\theta(C_t, t) \|_2^2
\bigr]
$.
This formulation enables the model to leverage both known-region context and outpainted-region cues while learning the reverse diffusion process for denoising.

\subsection{Inference}
At inference, M3DDM+ follows the M3DDM protocol: video outpainting is performed by applying the diffusion process over an extended canvas that spatially encompasses the input video. Because the VAE encoder is pretrained on $256 \times 256$ images, the model constructs a square ($1{:}1$ aspect ratio) latent representation, with the input video centered within this canvas, as illustrated in Fig.~\ref{fig:M3DDM_failure}.

If the extended region is initialized to zero, a sharp boundary appears between the known and extended regions. To address this, M3DDM applies OpenCV's image inpainting  \cite{telea_2004_opencv_inpainting} to smoothly fill the extended region.
At inference, the 3D U-Net takes as input a concatenated tensor
$C_t = \mathrm{concat}(Z_t, \tilde{Z}, m)$,
where $Z_t$ is the noised latent obtained by encoding the extended canvas with inpainting, $\tilde{Z}$ is the latent encoding of the masked version (that is, the extended canvas without inpainting), and $m$ is the downsampled binary mask of the extended region.

The latent video $\hat{Z}_0 = \{\hat{z}_{0,\tau}\}_{\tau=1}^{T}$ obtained from the reverse diffusion process is decoded frame-by-frame through the VAE decoder $D$ to produce
$\hat{x}_\tau = D(\hat{z}_{0,\tau})$.
The final output frame is then reconstructed by
$y_\tau = x_\tau \odot (1 - M_\tau) + \hat{x}_\tau \odot M_\tau$; that is,
blending the original input with the generated content according to the mask $M_\tau$,
preserving the known regions of $x_\tau$ while replacing the masked regions with the model's prediction $\hat{x}_\tau$.
Because the VAE operates on $256 \times 256$ inputs, inference is performed on a square ($1{:}1$) canvas that spatially encompasses the target aspect ratio. The final output is obtained by cropping the square canvas to the desired aspect ratio.

To leverage global temporal context, M3DDM uniformly samples frames from the entire masked video to form a set of global frames. These frames are encoded via the VAE encoder into latent representations, then projected by a lightweight learnable encoder to expand the channel dimension from 4 to 320, and fed as key/value inputs to the cross-attention layers of the 3D U-Net. The lightweight encoder is a shallow CNN trained jointly with the 3D U-Net to enable efficient cross-frame conditioning.

\section{EXPERIMENTAL RESULTS}

\subsection{Experimental Setup}

\paragraph{Benchmark and baseline.}
To evaluate the performance of M3DDM+, we perform experiments on the DAVIS dataset \cite{Perazzi_CVPR2016_DAVIS}, which contains 50 videos with an average length of 69.1 frames.
To isolate the impact of temporal information, we construct a dataset called \emph{DAVIS Static} by replacing all frames in each video with its first frame, enabling controlled comparison under no camera motion conditions.
We adopt M3DDM as our baseline, and since its training code is not publicly available, we evaluate M3DDM using the pretrained weights released on Hugging Face \cite{alimama_HuggingFace2024_M3DDM}.

\paragraph{Evaluation metrics.}
We evaluate our method using five standard video outpainting metrics: Mean Squared Error (MSE), Peak Signal-to-Noise Ratio (PSNR), Structural Similarity Index Measure (SSIM)\cite{Wang_IEEE2004_SSIM}, Learned Perceptual Image Patch Similarity (LPIPS)\cite{Zhang_CVPR2018_LPIPS}, and Fréchet Video Distance (FVD)\cite{unterthiner_arXiv2019_FVD}. MSE, PSNR, and SSIM primarily measure pixel-level distortion, while LPIPS and FVD assess perceptual quality and temporal coherence. A limitation of MSE is that it favors blurred outputs that match the color distribution of the ground truth over sharper outputs with different content, even when the latter are perceptually superior. 

To quantify the degree of blur in the outpainted region, we introduce a novel metric, Blurred Mean Squared Error (BMSE), which serves as a complementary metric to explicitly measure blur degradation in M3DDM and M3DDM+ outputs, as sharp generation is generally preferred. BMSE is computed by applying Gaussian blur to the generated video and measuring the MSE between the blurred and original versions. Sharp images exhibit larger changes under blurring (higher MSE), whereas already-blurred images change less. Therefore, lower BMSE values indicate a higher blur in the generated video.

\paragraph{Implementation details.}
M3DDM adapts the Stable Diffusion v1-5 \cite{Rombach_CVPR2022_SDv1-5} backbone to a 3D U-Net architecture, which incurs substantial computational costs for training from scratch on video generation tasks. To mitigate this, we fine-tune the publicly available pretrained M3DDM weights.
Following M3DDM, we train on a subset of the WebVid dataset \cite{Bain_ICCV2021_WebVid}, randomly selecting 10,000 videos to fine-tune.
We employ the Adam optimizer \cite{kingma_ICLR2015_adam} with a learning rate of 1e-5.
Both the local sliding window and the global context frames are set to 16 frames per forward pass.
For inference, we adopt a coarse-to-fine hierarchy with intervals [5, 3, 1] frames, tailored to the average sequence length of 69.1 frames in DAVIS.
During training, we adjust the masking strategy of M3DDM by setting the probabilities for 4-directional, 2-directional (vertical or horizontal) and 1-directional masks to 0.3, 0.55, and 0.15, respectively. The mask ratio is uniformly sampled from [0.15, 0.75], consistent with M3DDM.
For evaluation, we follow the M3DDM protocol and apply horizontal masking with ratios of 25\% and 66\% on both left and right sides of the input video, generating content in the masked regions.

\subsection{Results}

\paragraph{Quantitative results.}
As shown in Table~\ref{tab:quantitative}, our method achieves substantial improvements across multiple metrics compared to M3DDM. Although M3DDM attains marginally better pixel-level scores on DAVIS Static at 25\% mask ratio --- likely due to its reliance on temporal cues from adjacent frames --- its performance deteriorates as the mask ratio increases to 66\%. Our method consistently achieves higher BMSE scores in most settings, indicating a more effective mitigation of blur artifacts in generated regions.

\begin{table}[t]
    \centering

    \caption{Quantitative comparison of M3DDM and our M3DDM+ on DAVIS and DAVIS Static datasets at mask ratios of 25\% and 66\%. Our method consistently outperforms the baseline across distortion metrics (MSE, PSNR, SSIM), perceptual quality measures (LPIPS, FVD), and blur quantification (BMSE), with particularly significant improvements in information-limited scenarios with high mask ratios and no camera motion.}
    \label{tab:quantitative}

\vspace{.5em}

\resizebox{0.85\linewidth}{!}{
    \begin{tabular}{c|c|c|ccccc|c}
    dataset      & mask ratio & model & MSE $\uparrow$  & PSNR $\uparrow$  & SSIM $\uparrow$   & LPIPS $\downarrow$ & FVD $\downarrow$ & BMSE $\uparrow$   \\ \hline
    \multirow{4}{*}{DAVIS}        & \multirow{2}{*}{25\%}       & M3DDM & 0.00594          & 23.22          & 0.7815          & 0.1182          & 575           & 0.01931          \\
    \textbf{}    & \textbf{}  & Ours  & \textbf{0.00522} & \textbf{23.63} & \textbf{0.7889} & \textbf{0.0748} & \textbf{445}  & \textbf{0.02036} \\ \cline{2-9} 
                 &  \multirow{2}{*}{66\%}       & M3DDM & 0.02585          & 16.57          & 0.4964          & 0.4365          & 1426          & 0.01433          \\
                 &            & Ours  & \textbf{0.02092} & \textbf{17.64} & \textbf{0.5090}  & \textbf{0.3164} & \textbf{936}  & \textbf{0.01538} \\ \hline
     \multirow{4}{*}{DAVIS Static} &  \multirow{2}{*}{25\%}       & M3DDM & \textbf{0.05334} & \textbf{13.79} & \textbf{0.2267} & 0.4785          & 3333          & 0.01690           \\
                 &            & Ours  & 0.05505          & 13.65          & 0.2209          & \textbf{0.4422} & \textbf{3073} & \textbf{0.01816} \\ \cline{2-9} 
                 &  \multirow{2}{*}{66\%}       & M3DDM & 0.05976          & 12.96          & 0.2096          & 0.6003          & 3889          & \textbf{0.01449} \\
                 &            & Ours  & \textbf{0.05756} & \textbf{13.25} & \textbf{0.2132} & \textbf{0.5484} & \textbf{3442} & 0.01340          
    \end{tabular}
}

\end{table}

\begin{figure}[t]
    \centering
    \includegraphics[width=1\linewidth]{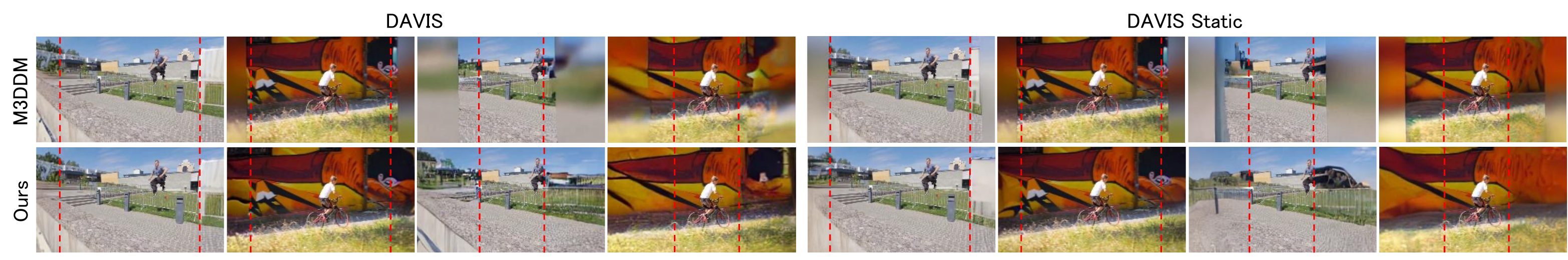}

    \caption{Qualitative comparison between M3DDM and M3DDM+. The red dotted line delineates the original video region (inside) from the outpainted region (outside). Our method effectively mitigates blur artifacts in challenging scenarios: static sequences with no camera motion (DAVIS Static) and cases requiring large mask ratios.}

    \label{fig:qualitative}
\end{figure}

\paragraph{Qualitative results.}
Fig.~\ref{fig:qualitative} presents a qualitative comparison. Our method substantially reduces blur artifacts in both challenging scenarios: sequences with no camera motion and cases requiring large mask ratios.

\section{CONCLUSION}
In this paper, we proposed M3DDM+, a refined training methodology that applies consistent mask processing across all frames to improve the generation quality of M3DDM. Our approach effectively mitigates blur artifacts and enhances visual fidelity in challenging scenarios where reference information is limited, including static scenes and large outpainting mask ratios. Quantitative and qualitative evaluations demonstrate that M3DDM+ achieves substantial improvements in perceptual quality while maintaining computational efficiency.

\acknowledgments
This work was supported in part by JSPS KAKENHI Grant Number JP22K12090 and 25K03138.

\bibliography{report}
\bibliographystyle{spiebib}

\end{document}